\documentclass[runningheads]{llncs}
\usepackage{graphicx}
\usepackage{comment}
\usepackage{amsmath,amssymb} 
\usepackage{color}
\usepackage{multirow}
\usepackage[colorlinks,linkcolor=black, pagebackref=false, anchorcolor=black,   citecolor=black]{hyperref}
\usepackage[misc]{ifsym}

\begin{document}
\pagestyle{headings}
\mainmatter

\title{Prime-Aware Adaptive Distillation} 

\titlerunning{Prime-Aware Adaptive Distillation}
%
\author{Youcai Zhang\thanks{This work was done when Youcai worked at Megvii Inc. Research Shanghai.} \and Zhonghao Lan\inst{2} \and Yuchen Dai\inst{1} \and Fangao Zeng\inst{1} \and Yan Bai\inst{3} \and Jie Chang\inst{1} \and Yichen Wei\inst{1}\textsuperscript{\Letter}}
\authorrunning{Y. et al.}
%

\institute{Megvii Inc. 
\email{\{daiyuchen,zengfangao,changjie,weiyichen\}@megvii.com}
\and University of Science and Technology of China, \email{lans@mail.ustc.edu.cn}\\
\and Tongji University, \email{yan.bai@tongji.edu.cn}\\
\email{yczhang12@fudan.edu.cn}
}
\maketitle

\begin{abstract}
Knowledge distillation(KD) aims to improve the performance of a student network by mimicing the knowledge from a powerful teacher network. Existing methods focus on studying what knowledge should be transferred and treat all samples equally during training. This paper introduces the adaptive sample weighting to KD. We discover that previous effective hard mining methods are not appropriate for distillation. Furthermore, we propose Prime-Aware Adaptive Distillation (PAD) by the incorporation of uncertainty learning. PAD perceives the prime samples in distillation and then emphasizes their effect adaptively. PAD is fundamentally different from and would refine existing methods with the innovative view of unequal training. For this reason, PAD is versatile and has been applied in various tasks including classification, metric learning, and object detection. With ten teacher-student combinations on six datasets, PAD promotes the performance of existing distillation methods and outperforms recent state-of-the-art methods.
\end{abstract}

\section{Introduction}
Obtaining highly accurate and lightweight (small model size and low computation) deep neural networks is crucial for practical applications. Knowledge Distillation (KD) methods~\cite{hinton2015distilling,romero2014fitnets,zagoruyko2016paying,park2019relational,tian2019contrastive,cho2019efficacy} are effective for this purpose and have been extensively explored in the past few years. KD aims to transfer the knowledge from a high-capacity \textit{teacher} network with a huge amount of parameters and heavy computation to a relatively lightweight \textit{student} network. By mimicking the ``behavior'' of teacher, the student model achieves better performance than as is trained from scratch.

Most knowledge distillation methods focus on studying \emph{what knowledge} should be transferred, {\it{e.g.}}, soft predictions~\cite{hinton2015distilling}, feature maps~\cite{romero2014fitnets}, activation-based attention maps~\cite{zagoruyko2016paying} or relation between samples~\cite{park2019relational}, etc. However, we notice that in the current literature it is rarely studied that \emph{which samples} contribute more to learning the student model and should be focused. 

This work, for the first time, studies the problem of adaptive sample weighting in knowledge distillation. First, we note that sample weighting has been widely used in other tasks, such as object detection~\cite{li2019gradient,lin2017focal} and metric learning~\cite{hermans2017defense,yu2018hard}. In these tasks, hard samples are assigned larger weights during training as they are more important to learn the model. These samples are termed ``prime samples'' in a recent work~\cite{cao2019prime}. However, such ``hard mining'' approaches are not appropriate in knowledge distillation. 
As validated in experiments, we found that hard samples have detrimental effect on learning the student model. We conjecture that the mismatch between student's and teacher's capacities makes student less capable of learning teacher's knowledge in these hard samples. Thus, we propose that sample weighting should be biased towards these easy samples.

Inspired by previous sample weighting approaches, we first introduce a few baseline 
weighting methods. They allocate weights based on the pre-defined function of sample's loss contribution. Samples with small loss are given large weights. However, these methods suffer from the sensitive hyper-parameters and the poor ability to distinguish prime samples.

To overcome the drawbacks, we propose a Prime-aware Adaptive Distillation (PAD) method. By modelling knowledge distillation with data uncertainty, each sample is modeled by a Gaussian distribution. The mean of the distribution is the prediction from the student and the variance measures the uncertainty about distilling the knowledge from the teacher. According to the estimated uncertainty from the network, PAD perceives the prime samples corrupted with weak noise in a more feasible and reasonable manner than the baseline methods.

In principle, our approach is fundamentally different from and would complement most (if not all) distillation methods. For this reason, PAD is a versatile method. It is applied in various tasks including classification, metric learning, and detection. It is validated to outperform our baseline weighting methods, and boost the performance of previous distillation methods, achieving new state-of-the-art methods on six datasets.

To summarize, this work makes three contributions.
\begin{enumerate} 
\item This is the first systematic study on the problem of sample weighting for knowledge distillation. In particular, we point out that previous hard mining approaches are not appropriate for our problem.  
\item We propose a simple yet effective distillation method (PAD). PAD helps the student network to perceive the prime samples in distillation by modelling the knowledge distillation with data uncertainty.
\item Comprehensive experiments validate that our approach is effective and establishes the new state-of-the-arts, on a vast range of datasets, baselines and previous distillation methods.
\end{enumerate}

\section{Related Work}
\subsubsection{Knowledge Distillation(KD).} Hinton et al.~\cite{hinton2015distilling} are pioneers in the field of knowledge distillation. They adopted the KL-divergence to penalize the softer probability distribution over classes between the teacher and student. From then on, many seminal studies~\cite{romero2014fitnets,zagoruyko2016paying,cho2019efficacy,tian2019contrastive,yuan2019revisit} have sprung up. For example, an information-theoretic framework~\cite{ahn2019variational} was proposed by maximizing the mutual information between the teacher and the student networks. The correlation between instances was also proved as the valuable knowledge by~\cite{park2019relational,peng2019correlation}. 

Besides studies on classification task, DarkRank~\cite{chen2018darkrank} was proposed to perform distillation in metric learning via cross sample similarities transfer. ROI mimic distillation~\cite{li2017mimicking} and fine-grained feature imitation~\cite{wang2019distilling} were designed for detection tasks by distilling the regions of interest.

\subsubsection{Adaptive Sample Weighting.}
Adaptive sample weighting by adjusting the contributions of samples for
training is a well-studied topic in computer vision. Hard-mining is a typical technique which plays a critical role in one-stage detectors~\cite{li2019gradient,lin2017focal} and metric learning~\cite{hermans2017defense,yu2018hard}. Hard-mining methods reduce the relative loss for easy samples, putting more focus on hard ones. In contrast, Huber loss and smooth $l_{1}$ loss reduce the contribution of hard samples by down-weighting the loss of them. Many researchers\cite{hu2019noise,zhong2019unequal} extended this idea to face recognition for noisy-robust feature learning. Cao et al.~\cite{cao2019prime} paid more attention to prime samples to achieve high detection performance. Our work is the first exploration about adaptive sample weighting in distillation, to our best knowledge. 

\subsubsection{Uncertainty Estimation.}
Uncertainty estimation is an effective technique to promote the robustness and interpretability of discriminant deep neural networks~\cite{nix1994estimating,isobe2017deep,kendall2017uncertainties}. In deep uncertainty learning, there are mainly two types of uncertainty. Model uncertainty estimates the noise of the parameters in deep neural networks, and data uncertainty measures the noise caused by input data. Predictions of neural networks are unreliable when the input sample is out of the training distribution or corrupted by noise.  Uncertainty estimation has attracted much attention in computer vision tasks, {\it{e.g.}}, face recognition~\cite{shi2019probabilistic}, semantic segmentation~\cite{isobe2017deep,kendall2017uncertainties}, object detection~\cite{choi2019gaussian,kraus2019uncertainty} and person re-identification(re-ID)~\cite{yu2019robust}. This paper models knowledge distillation with uncertainty estimation to predict the prime samples in distillation.

\section{Introducing Sample Weighting to KD}
\subsection{A General Formulation of KD}\label{sec:revisiting}
Most of the distillation studies obey a paradigm, 
whether they are in the task of classification, metric learning or object detection. The paradigm is to make students regress to the knowledge obtained from the teachers. The knowledge can be soft predictions~\cite{hinton2015distilling}, embedding feature~\cite{li2017mimicking}, feature maps of intermediate layer~\cite{romero2014fitnets}, or activation-based attention maps~\cite{zagoruyko2016paying}. Given the input $x_{i}\in X$, the regression aims to find the approximation function $f_{s}(\cdot)$ for student, where $\boldsymbol{f}_{s}(x_{i})$ should be close to the knowledge $\boldsymbol{y}_{i}\in Y$ derived from the teacher. 

A typical distillation loss can be formulated as

\begin{equation}
    \mathcal{L}_{distill} = \frac{1}{N}\sum_{i=1}^{N}w_{i}d[\boldsymbol{f}_{s}(x_{i}), \boldsymbol{y}_{i}].
\end{equation}
$N$ is the number of samples in a batch. $d[\cdot,\cdot]$ is the distance metric to measure the gap between the student and the teacher. $L_{2}$ is in common use as the metric. If there is a mismatch of dimension between $\boldsymbol{f}_{s}(x_{i})$ and $\boldsymbol{y}_{i}$, a simple projection function will be added for dimension conversion, following FitNet~\cite{romero2014fitnets}. For notation convenience, we abbreviate the gap $d[\boldsymbol{f}_{s}(x_{i}), \boldsymbol{y}_{i}]$ between teacher and student for sample $x_{i}$ to $d(x_{i})$. Noted $w_{i}$ is the weight to measure the importance of $x_{i}$ to the overall loss. $w_{i}$ is set to be same for all samples in conventional knowledge distillation methods. 

To guide the learning of student network, distillation loss should be combined with the original task-specific loss $\mathcal{L}_{task}$. The overall loss is formulated as

\begin{equation}
    \mathcal{L} = \mathcal{L}_{task} + \lambda \mathcal{L}_{distill},
\end{equation}
where $\lambda$ is a hyper-parameter to balance the task loss and distillation loss.

\subsection{Sample Weighting and A Few Baselines}\label{sec:manualways}
Conventional distillation methods treat all samples equally, without considering the difference among them. Hard sample mining is a well-studied sample weighting method in metric learning~\cite{yu2018hard} and object detection~\cite{lin2017focal}. The loss contribution usually reflects the level of sample difficulty~\cite{lin2017focal,yu2018hard}. The greater the loss is, the more difficult the sample is. Therefore, weighting samples according to their loss contribution is common practice in hard example mining. 

Following~\cite{yu2018hard}, we also conduct hard sample mining by increasing the weights of samples with large gaps($d$) in distillation. Table~\ref{table:hard-mi} shows hard-mining is conversely worse than treating samples equally in distillation. Thus, we argue that hard samples are not prime ones in distillation. For further verification, we design a simple weighting scheme, discarding hard samples with the largest distillation gaps($d$) in each batch during training. We find that training without the hardest samples slightly improves the baselines. Combining the hard-mining and hard-discarding results, we can come to the conclusion that hard samples have detrimental effect on the training of the student model.

\begin{table}
\begin{center}
\caption{Comparisons of hard-mining (training with large weights for hard samples) and hard-discarding (training without the hardest samples). 
``$\downarrow$'' means worse than baseline. Experimental settings will be introduced in detail in Sec.~\ref{sec:exp}.}
\label{table:hard-mi}
\begin{tabular}{c|c|c|c|c|c}
\hline
Method                     & CIFAR100 & TinyImageNet & Duke-reID  & CUB-200 & Pascal VOC07\\ \hline
all samples equally        & 72.85    & 63.08   & 79.5          & 58.6  &45.53       \\ \hline
hard-mining                & 72.12$\downarrow$    & 62.57$\downarrow$   & 79.0$\downarrow$          & 55.7$\downarrow$  & 45.44$\downarrow$ \\
hard-discarding               & 72.91    & 63.16   & 79.8          & 58.3$\downarrow$  & 45.94 \\ \hline
\end{tabular}
\end{center}
\end{table}

Besides the above simple either-or method, we continue to design better weighting schemes. The weighting function $w$ can be formulated as $w_{i} = w(d(x_{i}))$. $w(\cdot)$ is a monotone increasing function instead of a decreasing one in hard example mining. We design two soft weighting schemes, exponential weighting and polynomial weighting~\cite{yu2018hard}, to allocate each sample a continuous weight. Exponential weighting is defined as
\begin{equation}
     w_{\text{\it{soft-exp}}}(d(x_{i})) = \frac{e^{-d(x_{i})/T}}{\sum_{i=1}^{N}e^{-d(x_{i})/T}}, 
\end{equation}
where $T>0$ controls the power of the suppression against the outliers. And polynomial weighting function can be formulated as
\begin{equation}
    w_{\text{\it{soft-poly}}}(d(x_{i})) = \frac{(1+d(x_{i}))^{-\alpha}}{\sum_{i=1}^{N}(1+d(x_{i}))^{-\alpha}},
\end{equation}
where $\alpha > 0$ is a coefficient for adjusting the weight distribution.

\section{Prime-Aware Adaptive Distillation}
Besides easy sample mining in distillation, there also exist problems in introducing sample weighting to distillation. The performance of the above baselines is not satisfactory and sensitive to
the additional hyper-parameters as shown in Table~\ref{table:ablation}.
In light of this, we propose a novel Prime-aware Adaptive Distillation (PAD) method by modeling knowledge distillation with data uncertainty. As data uncertainty can capture the noise inherent in the observations, PAD can perceive the prime samples with little noise and weight them adaptively. 

\begin{figure}
\centering
\includegraphics[height=3.8cm]{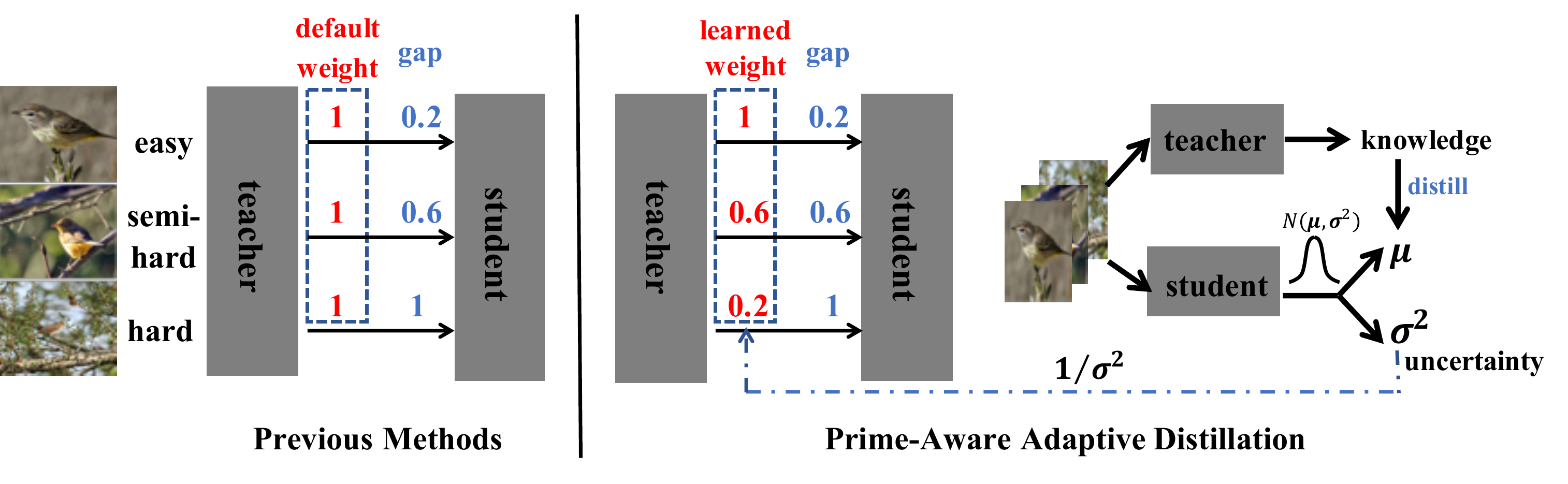}
\caption{Previous methods ignore the quality of different samples and allocate the same weight to them. The proposed method perceives the prime samples (usually with small gaps) and enhances them by given large weights, {\it{i.e.}}, Prime-aware Adaptive Distillation (PAD). Each sample is modeled by a Gaussian distribution $N(\mu,\sigma^{2})$. The mean $\mu$ is the prediction of student and the variance $\sigma^{2}$ measures the uncertainty about distilling the knowledge from the teacher. Prime samples are assigned large weights ($1/\sigma^{2}$), {\it{i.e.}}, small variance ($\sigma^{2}$), indicating that the student is confident about their knowledge.}
\label{fig:intro}
\end{figure}

\subsubsection{Modelling KD with data uncertainty.}
As discussed in Sec.~\ref{sec:revisiting}, KD can be viewed as a regression task. In most real-world scenarios, paired input-target data contains noise. The target values $\boldsymbol{f}_{s}(x_{i})$ is inevitably corrupted by input-dependent noise $\boldsymbol{n}(x_{i})$. The observed data can be modeled by $\boldsymbol{y}_{i} = \boldsymbol{f}_{s}(x_{i}) + \boldsymbol{n}(x_{i})$, where the additive noise $\boldsymbol{n}(x_{i})$ can be viewed as errors that move the targets away from their true values $\boldsymbol{f}_{s}(x_{i})$ to their observed values $\boldsymbol{y}_{i}$. 

Supposing the noise obeys Gaussian distribution as $\boldsymbol{n}(x_{i})\sim N(0, \boldsymbol{\sigma}^2(x_{i}))$, where $\boldsymbol{\sigma}^2$ is the variance of the noise. The target probability distribution $\boldsymbol{y}_{i}$ for input ${x}_{i}$ based on the least-square regression can be formulated as
\begin{equation}
  p(\boldsymbol{y}_{i}|x_{i}) =  \frac{1}{\sqrt{2 \pi \boldsymbol{\sigma}_{i}^{2}}} \exp \left(-\frac{\left(\boldsymbol{f}_{s}(x_{i}) - \boldsymbol{y}_{i}\right )^{2}}{2 \boldsymbol{\sigma}_{i}^{2}}\right),
\end{equation}
where $\boldsymbol{f}_{s}(x_{i})$ corresponds to the mean $\boldsymbol{\mu}_{i}$ of this distribution and $\boldsymbol{\sigma}^2_{i}$ is the variance to measure the uncertainty of the predicted value $\boldsymbol{f}_{s}(x_{i})$. 

We should maximize the log likelihood for the input $\boldsymbol{x}_{i}$ over the observation $\boldsymbol{y}_{i}$, so the negative log likelihood is formulated as
\begin{equation}
-\ln p\left(\boldsymbol{y}_{i} | x_{i}\right)=\frac{\left(\boldsymbol{f}_{s}(x_{i}) - \boldsymbol{y}_{i}\right)^{2}}{2 \boldsymbol{\sigma}_{i}^{2}} + \frac{1}{2} \ln \boldsymbol{\sigma}_{i}^{2} + \frac{1}{2} \ln 2 \pi.
\end{equation}

Based on the above derivation, we can formulate a prime-aware distillation loss to help
the student network perceive sample difficulty, as:
\begin{equation}
    \mathcal{L}_{PAD} = \sum_{i=1}^{N}(\frac{(\boldsymbol{f}_{s}(x_{i})-\boldsymbol{y}_{i})^{2}}{\boldsymbol{\sigma}_{i}^{2}}+\ln \boldsymbol{\sigma}_{i}^{2}).
    \label{eq:PAD}
\end{equation}
Without loss of generality, we take the embedding feature as the distillation target for simplicity. It is easy to generalize to other forms of targets, {\it{e.g.}}, by flattening feature map to a vector, we can treat feature map as the feature. Due to the least-square assumption, Eq.~(\ref{eq:PAD}) can be directly applied to $L_{2}$-based methods, {\it{e.g.}}, $L_{2}$ on the feature, FitNet~\cite{romero2014fitnets}, AT~\cite{zagoruyko2016paying}, and many detection distillation methods~\cite{li2017mimicking,wang2019distilling}. 

For the application of PAD, the only modification is the addition of a variance branch parallel with $\boldsymbol{f}_{s}(x)$ to estimate the variance $\boldsymbol{\sigma}$. And $\boldsymbol{\mu}$ can be viewed as a drop-in replacement of $\boldsymbol{f}_{s}(x)$ as shown in Fig.~\ref{fig:intro}. Using an auxiliary branch to estimate the variance is a common practice and has been proved to be effective in data uncertainty~\cite{nix1994estimating,isobe2017deep,kendall2017uncertainties,yu2019robust}. The motivation of an auxiliary branch is its provision of the network with the ability to measure the uncertainty. It should also be pointed out that the variance branch only exists in training, thus it will not bring about any extra computational cost for inference. We also find that simple design for the variance branch achieves satisfactory results and the variance branch has almost no effect on training efficiency.

\subsubsection{Discussion.} 
In this part, we analyze how uncertainty effects the distillation from the perspective of loss function. First, Eq.~(\ref{eq:PAD}) indicates the effect from samples with large variances is weakened in network training. And then, a question comes out: what kind of sample will be given a large variance? $ln(\boldsymbol{\sigma}^2)$ can be regarded as a regularization term, which prevents the model predicting large variances for all samples. Simultaneously, model will not produce small variances for all samples, which will make the first term of Eq.~(\ref{eq:PAD}) increase rapidly. The model therefore allocates large variances to samples with large gaps($d$) to reduce the overall loss of Eq.~(\ref{eq:PAD}).
Samples with large gaps($d$) are usually difficult for student. That explains why they are termed hard samples. The rationality of the above hypothesis is proved by experimental analysis in Sec.~\ref{sec:exp_analysis}.

To sum up, $\frac{1}{\sigma^{2}}$ can be viewed as a weight coefficient. PAD actually provides a mechanism of adaptive weighting different training samples. Easy samples are highlighted and hard samples are restricted under the influence of uncertainty. 

\subsubsection{Comparison with sample weighting baselines.} PAD obtains the weight directly from the network instead of a conversion from the distance by a pre-defined function. The weights learned from PAD are more accurate in describing sample difficulty than those from baselines, which can be proved by Table~\ref{table:diff scheme} and Fig.~\ref{fig:var_vis}. Besides, PAD introduces no extra hyper-parameters that need to be tuned carefully in those baseline methods. For this reason, PAD exhibits a versatile generalization on various methods and tasks.

\subsubsection{Comparison with VID~\cite{ahn2019variational}.} Though the formula of VID and PAD is similar, the difference is minor but essential. VID is one of the existing distillation works which treats samples equally. As a contrast, PAD is motivated by the "adaptive sample weighting" ({\it{i.e.}}, unequal training), which is well-studied in CV community but missing in distillation. Our study is the first work to exploit adaptive sample weighting in distillation and proposes PAD as an effective weighting way. This difference between PAD and VID results in the different derivation of loss and the meaning of variance. PAD is derived from the data uncertainty, while VID is formulated by the information theory. Also, the variance of PAD has a clear meaning, which indicates the strong correlation between variance and sample quality. Last but not least, PAD performs better than VID with a remarkable margin as shown in Table \ref{table:cifar}.

\section{Experiment}\label{sec:exp}
In this section, we firstly apply the proposed PAD on three visual tasks: image classification, metric learning, and object detection. Our experiments exhibit two advantages over others. 1) {\textbf{Stronger baselines}}: student baselines in our experiments are obviously higher than those in recent studies~\cite{park2019relational,tian2019contrastive,wang2019distilling}, which will be elaborated in the following part. 2) {\textbf{More extensive experiments on more challenging tasks}}: we conduct experiments with ten teacher-student combinations on various tasks, including ImageNet, where it is hard to achieve positive results for distillation~\cite{zagoruyko2016paying}. As a comparison, RKD~\cite{park2019relational} uses five teacher-student combinations and lacks of experiments on large-scale datasets. Also, we compare different methods on detection, whereas almost all the related works~\cite{wang2019distilling,li2017mimicking,gao2018embarrassingly,chen2017learning} only report their methods without any comparison. 

Secondly, comprehensive analysis is given to delve into the learning process of PAD. Finally, we conduct ablation experiments to compare different unequal-learning schemes. All the experiments are conducted on Pytorch. For all the datasets, we follow the train/test splits suggested as popular papers. We carefully tune the weight of distillation loss $\lambda$ and use the SGD optimizer with a momentum of 0.9 for all the experiments.

\subsection{Classification}
Firstly, we evaluate PAD on the image classification task where most knowledge distillation methods report their performance.

\noindent \textbf{Dataset settings.} CIFAR100~\cite{krizhevsky2009learning}, TinyImageNet~\cite{le2015tiny}, and ImageNet~\cite{deng2009imagenet} are adopted. CIFAR100 contains 50K training images with 500 images per class and 10K test images. TinyImageNet has 200 classes, each with 500 training images and 50 validation images. ImageNet~\cite{deng2009imagenet} provides 1.2 million images from 1K classes for training and 50K for validation.

\noindent \textbf{Implementation details.} Following original papers or popular implementations, we apply $L_{2}$, RKD~\cite{park2019relational}, and CC~\cite{peng2019correlation} on the last embedding layer before classification, apply the FitNet~\cite{ba2014deep} on the last two blocks of CNN, and apply AT~\cite{zagoruyko2016paying} and VID~\cite{ahn2019variational} on the last four blocks. We re-implement the HKD~\cite{hinton2015distilling}, FitNet~\cite{ba2014deep}, and $L_{2}$ based on the original paper. For RKD, CC, and AT, we use author-provided codes. For VID, we use the author-verified code from~\cite{tian2019contrastive}.

For CIFAR100 and TinyImageNet, we run each model for 200 epochs with a batch size of 128, and the weight decay of 5e-4. We set the initial learning rate to 0.1, dropping 0.2x at 60, 120, 160 epoch. For ImageNet, we run each model for 90 epochs with a batch size of 512, and set the initial learning rate to 0.2, dropping 0.1x at 30, 60 epoch. The weight decay is set to 1e-4. 
When combined with $L_{2}$ and AT, PAD adopts a fully connected(FC) layer, followed by a batch normalization(BN) layer to generate variance. When combined with FitNet, PAD adopts a $1\times1$ convolutional layer followed by a BN layer.
\subsubsection{Results on CIFAR100 and TinyImageNet.}
We adopt the ResNet18~\cite{he2016deep} as the teacher network, MobileNetV2~\cite{sandler2018mobilenetv2} as the student network. We report the top-1 test accuracy on CIFAR100 and TinyImageNet datasets in Table~\ref{table:cifar}.
\begin{table}
\begin{center}
\caption{Top-1 accuracy (\%) on CIFAR100 and TinyImageNet}
\label{table:cifar}
\resizebox{\textwidth}{!}{
\begin{tabular}{c|cc|cccc|cccccc}
\hline
             & teacher & student & HKD & RKD & CC & VID & AT & PAD-AT         & $L_{2}$ & PAD-$L_{2}$   & FitNet & PAD-FitNet     \\ \hline
CIFAR100    & 75.86   & 68.16   & 70.29    & 68.34     & 70.0     & 68.2      & 69.06    & \textbf{69.92} & 72.85          & \textbf{74.06} & 71.74        & \textbf{73.45} \\
tinyImageNet & 63.46   & 56.16   & 59.52    & 55.88     & 57.14    & 57.06     & 58.24    & \textbf{59.26} & 63.08          & \textbf{65.64} & 63.34        & \textbf{64.20} \\ \hline
\end{tabular}}
\end{center}
\end{table}

Results of baseline and compared methods by our implementation are reliable,  as they are almost the same as those of~\cite{yuan2019revisit}, and are higher than those of~\cite{tian2019contrastive}. Weobserve that PAD promotes the performance of three existing classic knowledge distillation methods(AT~\cite{zagoruyko2016paying}, $L_{2}$, and FitNet~\cite{romero2014fitnets}) with significant margins. For example, PAD obtains a 1.71\% gain for FitNet on CIFAR100 and 2.56\% for $L_{2}$ on TinyImagenet. PAD combined $L_{2}$ outperforms other methods. We also find that relation-level distillation methods (CC~\cite{peng2019correlation} and RKD~\cite{park2019relational}) are not superior to the instance-level methods (HKD~\cite{hinton2015distilling}, FitNet~\cite{romero2014fitnets}, and $L_{2}$) on these two classification tasks. VID~\cite{ahn2019variational} also performs worse than HKD~\cite{hinton2015distilling}. Similar observations can be found in~\cite{tian2019contrastive}. Compared with VID~\cite{ahn2019variational}, our method obtains significant improvement, which shows the effectiveness of allocating different weights to different samples instead of a same weight.

\subsubsection{Results on ImageNet.}
For fair comparisons with AT~\cite{zagoruyko2016paying} and CRD~\cite{tian2019contrastive}, we adopt the models from them, ResNet34 as the teacher and ResNet18 as the student. As shown in Table~\ref{table:imagenet}, $L_{2}$ performs better than FitNet and AT on ImageNet, so we just apply PAD to $L_{2}$. It is worth mentioning that $L_{2}$ on the last embedding layer is an effective method for distillation, but we do not find its previous use in classification. The gap of top-1 accuracy between the teacher and student is 3.56\%. Our PAD-$L_{2}$ method reduces this gap by 1.96\%, ahead of the state-of-the-art CRD with a margin of 0.54\%. Results on ImageNet indicate that our method is generically applicable in the large-scale classification task.

\begin{table}
\begin{center}
\caption{Top-1 accuracy (\%) on ImageNet. ``*'' means the result from the paper. }
\label{table:imagenet}
\begin{tabular}{cc|ccccc|cc}
\hline
teacher & student & HKD\cite{hinton2015distilling}       & FitNet\cite{romero2014fitnets}  & CC\cite{peng2019correlation}    & AT\cite{zagoruyko2016paying}    & CRD$^\star$\cite{tian2019contrastive}  & $L_{2}$ & PAD-$L_{2}$   \\ \hline
73.31   & 69.75   & 70.80    & 70.62   & 70.74 & 70.43 & 71.17  & 70.90         & \textbf{71.71} \\ \hline
\end{tabular}
\end{center}
\end{table}

\subsection{Metric Learning}
Secondly, we demonstrate the effectiveness of PAD on metric learning.

\noindent \textbf{Dataset settings.} We consider two typical metric learning tasks, {\it{{\it{i.e.}}}}, fine-grained image retrieval on CUB-200-2011~\cite{wah2011caltech} and person re-ID on DukeMTMC-reID~\cite{ristani2016performance}. CUB-200-2011 contains 200 different classes of birds. We use the first 100 classes with 5,864 images for training and the last 100 classes with 5,924 images for testing. $R@1$ is adopted as the evaluation metric. DukeMTMC-reID is a subset of the DukeMTMC dataset designed for person re-ID. It consists of 36,411 human bounding boxes belonging to 1,404 identities. The training set contains 16,522 images of 702 identities and the rest 702 identities are assigned to the testing set. $R@1$ and mean accuracy precision($mAP$) are adopted as metrics.

\noindent \textbf{Implementation details.} Embedding feature is directly used for retrieval in metric learning, thus, we choose the embedding feature as the distillation target layer for all methods. There are few works designed specifically for metric learning. We choose typical $L_{2}$, Darkrank~\cite{chen2018darkrank} and RKD~\cite{park2019relational} as the compared methods. Two losses are proposed in Darkrank~\cite{chen2018darkrank}, {\it{{\it{i.e.}}}}, HardRank and SoftRank loss. We adopt the HardRank loss as it is computationally efficient and also comparable to the SoftRank loss. For RKD~\cite{park2019relational}, we apply both RKD-D and RKD-A with a weight 1:2 on the feature without normalization. The paper~\cite{park2019relational} suggests that the student should be trained purely by the RKD loss, {\it{i.e.}}, removing the task loss. We report results of the student trained with and without task loss.

For CUB-200-2011, we train for 30,000 steps with a batch size of 60 (12 classes and each class 5 samples), weight decay of 4e-5. The initial learning rate is 1e-3, and is divided by 10 every 10,000 steps. For DukeMTMC-reID, the batch size is 36 (12 classes and each class 3 samples) and the initial learning is 0.005 decaying once at 20,000 step. Other settings are same to CUB-200-2011. The variance branch adopts a FC layer followed by BN.   

\subsubsection{Results on CUB-200-2011.}
Inspired by RKD, we conduct both compression distillation and self-distillation on CUB-200-2011 with a strong baseline. Compression distillation means distillation to a smaller network and self-distillation means the teacher and student share the same architecture. For the teacher, we adopt the GoogLeNet~\cite{szegedy2015going} trained by multi-similarity loss~\cite{wang2019multi}, which provides a higher $R@1$ of 64.7 than 61.2 in RKD paper. A strong teacher is beneficial to explore the real performance of self-distillation. For the student in compression distillation, we choose ResNet18~\cite{he2016deep} with multi-similarity loss, which also provides a higher $R@1$ of 55.6 than 53.9 in RKD paper.

Table~\ref{table:metriclearning} shows the results of different distillation methods on CUB-200-2011. RKD achieves promising results and PAD outperforms the Darkrank and RKD. Self-distillation further improves the performance of the initial teachers. Although our reproduced results (64.7) are inferior to those reported in the paper (65.7)~\cite{wang2019multi}, self-distillation by Darkrank, RKD and our PAD outperform the initial teacher and exceeded the reported results of 65.7. The proposed method brings a 1.6\% gain. Similarly to RKD, the repeated self-distillation does not provide additional benefits.

\begin{table}
\caption{Results on metric learning tasks. w/o task indicates students are trained without task loss, only by RKD. w/ task indicates students are trained with both task loss and RKD.}
\resizebox{\textwidth}{!}{
\begin{tabular}{ccc|cc|cc|cc}
\cline{2-9}
                                           & \multicolumn{6}{c|}{DukeMTMC-reID}                                                                                & \multicolumn{2}{c}{CUB-200-2011}                                  \\ \cline{2-9} 
                                           & \multicolumn{4}{c|}{compression distillation}                            & \multicolumn{2}{c|}{self-distillation} & \multicolumn{1}{c|}{compression distillation} & self-distillation \\ \cline{2-9} 
                                           & \multicolumn{2}{c|}{softmax+triplet} & \multicolumn{2}{c|}{only triplet} & \multicolumn{2}{c|}{softmax+triplet}   & \multicolumn{1}{c|}{multi-similarity}          & multi-similarity   \\ \hline
\multicolumn{1}{c|}{Method}                & $R@1$               & $mAP$              & $R@1$             & $mAP$             & $R@1$                & $mAP$               & \multicolumn{1}{c|}{$R@1$}                      & $R@1$               \\ \hline
\multicolumn{1}{c|}{teacher}               & 84.9              & 71.9             & 84.9            & 71.9            & 84.9               & 71.9              & \multicolumn{1}{c|}{64.7}                     & 64.7              \\
\multicolumn{1}{c|}{student}               & 79.3              & 61.1             & 63.6            & 43.5            & 84.9               & 71.9              & \multicolumn{1}{c|}{55.6}                     & 64.7              \\ \hline
\multicolumn{1}{c|}{DarkRank}              & 79.9              & 62.2             & 69.9            & 50.9            & 85.2               & 72.2              & \multicolumn{1}{c|}{60.1}                     & 65.2              \\
\multicolumn{1}{c|}{RKD w/ task}           & 68.3              & 48.9             & 68.3            & 48.9            & 73.1               & 55.3              & \multicolumn{1}{c|}{60.7}                     & 65.8              \\
\multicolumn{1}{c|}{RKD w/o task}          & 80.3              & 63.2             & 70.9            & 52.3            & 85.1               & 73.1              & \multicolumn{1}{c|}{60.7}                     & 63.0              \\
\multicolumn{1}{c|}{$L_{2}$}              & 79.5              & 62.5             & 72.6            & 53.7            & 86.0               & 72.8              & \multicolumn{1}{c|}{58.6}                     & 64.4              \\
\multicolumn{1}{c|}{PAD-$L_{2}$} & \textbf{81.0}     & \textbf{63.3}    & \textbf{77.5}   & \textbf{60.5}   & \textbf{87.3}      & \textbf{74.3}     & \multicolumn{1}{c|}{\textbf{61.4}}                     & \textbf{66.3}              \\ \hline
\end{tabular}}
\label{table:metriclearning}
\end{table}

\subsubsection{Results on DukeMTMC-reID.}
For person re-ID, we also choose a strong baseline. The teacher is the ResNet50 trained with both softmax and triplet loss, which gives a gain of 2.1\% $R@1$ than only softmax. Multi-similarity loss gives no additional gain than triplet loss in our experiments. Table~\ref{table:metriclearning} shows distillation results of different students with different methods. As we see, improvement obtained by distillation is slight when the student baseline is strong. While the improvement becomes remarkable, when the student is trained only by triplet loss. The proposed PAD outperforms RKD and DarkRank based on different baselines. Besides, PAD demonstrates excellent performance in self-distillation compared with other methods. 

\subsection{Object Detection}
Thirdly, we validate the proposed method on object detection which is a fundamental and more challenging task in computer vision.

\noindent \textbf{Dataset settings.} We conduct experiments on Pascal VOC dataset~\cite{everingham2010pascal} using both two-stage (Faster-RCNN~\cite{ren2015faster}) and one-stage (RetinaNet~\cite{lin2017focal}) frameworks. Following ROI Mimic~\cite{li2017mimicking}, we use VOC2007 trainval set of 5k images and VOC2012 trainval set of 16k images as training data. And we evaluate our method on VOC2007 test set of 5K images. Besides $mAP@0.5$ that considers one generous threshold of IoU $\geqslant$ 0.5, we adopt the overall $mmAP$ metric, averaging over the 10 IoU thresholds. 

\noindent \textbf{Implementation details.} All the detection experiments including baselines are implemented on Detectron2~\cite{wu2019detectron2}, which provides the strongest baselines for popular object detection frameworks. For example, Detectron2’s ResNet50 based Faster R-CNN achieves a $mAP@0.5$ of 80.9 while the fine-grained paper~\cite{wang2019distilling} reports that of only 69.1. All the hyper-parameters related to the object detection are consistent with the standard configurations provided by Detectron2. We only tune the parameters of the distillation part.

For Faster R-CNN, the variance branch consists of two 3x3 convolutional layers and two FC layers, generating variance for each ROI extracted by the sampler. For RetinaNet, variance branch consists of five convolutional layers for the whole feature map and then the spatial mask generated in Fine-frained's way~\cite{wang2019distilling} filters out the background. Since FPN is adopted in all experiments, distilled features are sampled from all FPN layers.

\subsubsection{Results on Pascal VOC07.}
Following fine-grained~\cite{wang2019distilling}, we adopt two settings of backbones, {\it{{\it{i.e.}}}}, from ResNet101 to ResNet50 and from VGG16 to VGG11. Table~\ref{table:detection} summaries the results of different distillation methods based on different architectures. We have two observations from Table~\ref{table:detection}. First, PAD is applied to the two most mainstream frameworks and consistently improves the performance of students based on different backbones. Second, the improvement brought by distillation is not as impressive as the original paper~\cite{li2017mimicking,wang2019distilling}. We argue that we should analyze the improvement from a relative rather than absolute perspective when baselines become strong. For example, in first column of Table~\ref{table:detection}, ROI mimic~\cite{li2017mimicking} improves the $mmAP$ of student from 54.00\% to 55.52\%. This gain covers 68\% of the gap between teacher and student. Our PAD not only further narrows the gap but also makes the student exceed the teacher slightly, which is a rare phenomenon in detection distillation. Similarly, Fine-grained method covers 54.40\% and 59.25\% of the $mmAP$ gaps using two backbones respectively based on RetinaNet. Combined with PAD, the gaps decrease by 90.67\% and 93.58\% and the performance of students become very close to those of teachers. Results using two typical distillation methods based on Faster-RCNN and RetinaNet show that PAD still performs well on detection.

\begin{table}
\caption{Results on Pascal VOC07 with different backbones. ResNet101-50 refers to the backbone of teacher and student respectively. Results of ROI-mimic based on RetinaNet are missing, as ROI-mimic can only be applied to two-stage frameworks.}
\resizebox{\textwidth}{!}{
\begin{tabular}{ccccc|cc|cc}
\cline{2-9}
                                                 & \multicolumn{4}{c|}{Faster R-CNN}                                                       & \multicolumn{4}{c}{RetinaNet}                                      \\ \cline{2-9} 
                                                 & \multicolumn{2}{c|}{ResNet101-50}                    & \multicolumn{2}{c|}{VGG16-11}    & \multicolumn{2}{c|}{ResNet101-50} & \multicolumn{2}{c}{VGG16-11}    \\ \hline
\multicolumn{1}{c|}{\textbf{Method}}             & \textbf{$mmAP$}  & \multicolumn{1}{c|}{$mAP@0.5$}        & \textbf{$mmAP$}  & $mAP@0.5$        & \textbf{$mmAP$}   & $mAP@0.5$         & \textbf{$mmAP$}  & $mAP@0.5$        \\ \hline
\multicolumn{1}{c|}{teacher}                     & 56.22          & \multicolumn{1}{c|}{82.53}          & 46.53          & 76.86          & 57.45           & 82.09           & 47.18          & 74.34          \\
\multicolumn{1}{c|}{student}                     & 54.00          & \multicolumn{1}{c|}{81.98}          & 42.18          & 72.99          & 55.52           & 81.21           & 43.13          & 69.57          \\ \hline
\multicolumn{1}{c|}{ROI mimic~\cite{li2017mimicking}}                   & 55.52          & \multicolumn{1}{c|}{82.25}          & 45.00          & 75.04          & -               & -               & -              & -              \\
\multicolumn{1}{c|}{PAD-ROI-mimic}    & \textbf{56.41} & \multicolumn{1}{c|}{\textbf{82.46}} & \textbf{45.62} & \textbf{75.79} & -               & -               & -              & -              \\
\multicolumn{1}{c|}{Fine-grained~\cite{wang2019distilling}}                & 54.91          & \multicolumn{1}{c|}{81.93}          & 44.60          & 74.59          & 56.57           & 81.46           & 45.53          & 71.99          \\
\multicolumn{1}{c|}{PAD-Fine-grained} & \textbf{55.39} & \multicolumn{1}{c|}{\textbf{82.29}} & \textbf{45.21} & \textbf{75.17} & \textbf{57.27}  & \textbf{81.94}  & \textbf{46.92} & \textbf{73.18} \\ \hline
\end{tabular}
}
\label{table:detection}
\end{table}

\subsection{Analysis}\label{sec:exp_analysis}
Finally, we give visualization analysis to show the mechanism and effectiveness of PAD. Also, we compare different weighting schemes and conduct ablation experiments. Baselines in this section are students distilled with $L_{2}$.

\begin{figure}
	\begin{minipage}[t]{0.5\linewidth}
		\centering
		\includegraphics[width=2.5in]{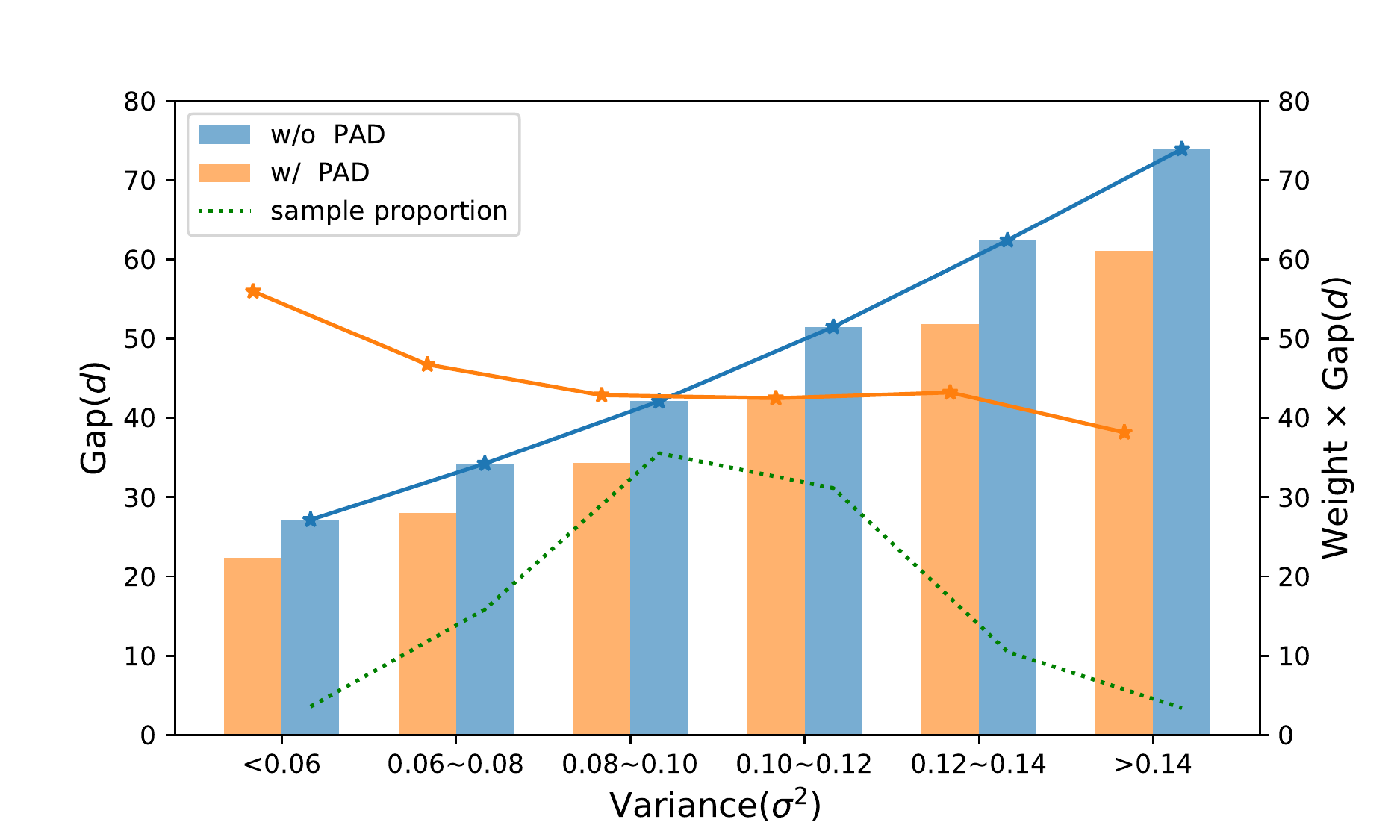}
	\end{minipage}
	\begin{minipage}[t]{0.5\linewidth}
		\centering
		\includegraphics[width=2.5in]{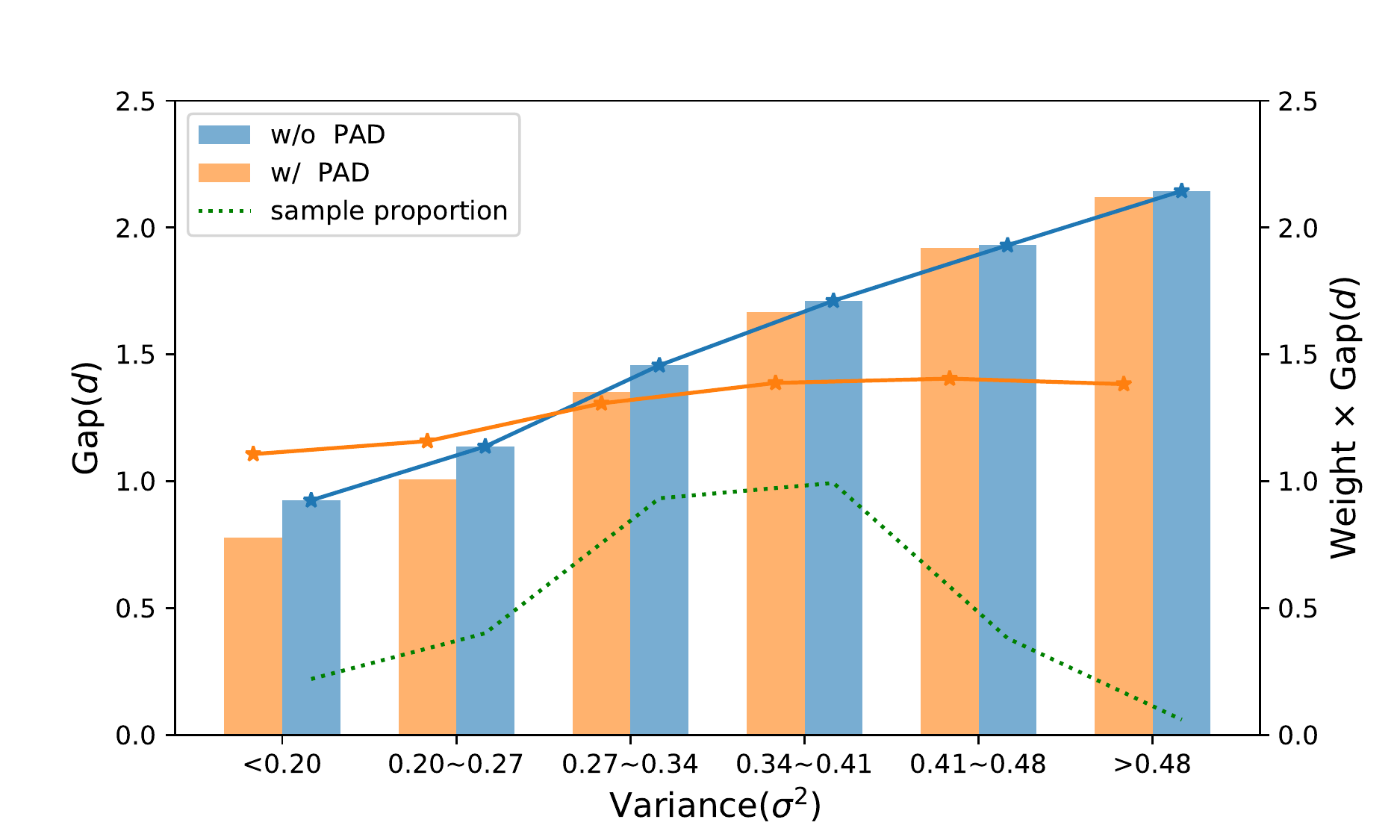}
	\end{minipage}

\caption{Relationship between the variance($\sigma^{2}$) and gap($d$) on the training set of TinyImageNet({$\it{left}$}) and CUB-200-2011({$\it{right}$}). Bars refer to the gap($d$) between teacher and student for different samples, and x-axis is the corresponding variance bins of these samples. Green dotted lines indicate the sample proportion. Solid lines demonstrate the actual effect ($Weight \times Gap$) of different samples on distillation. The effect of samples with large variances is obviously weakened. Note that the absolute effect of different methods should not be compared, as we tune the distillation loss weight $\lambda$ for all methods.}
\label{fig:var_gap}
\end{figure}

\begin{figure}
\centering
\includegraphics[height=4.3cm]{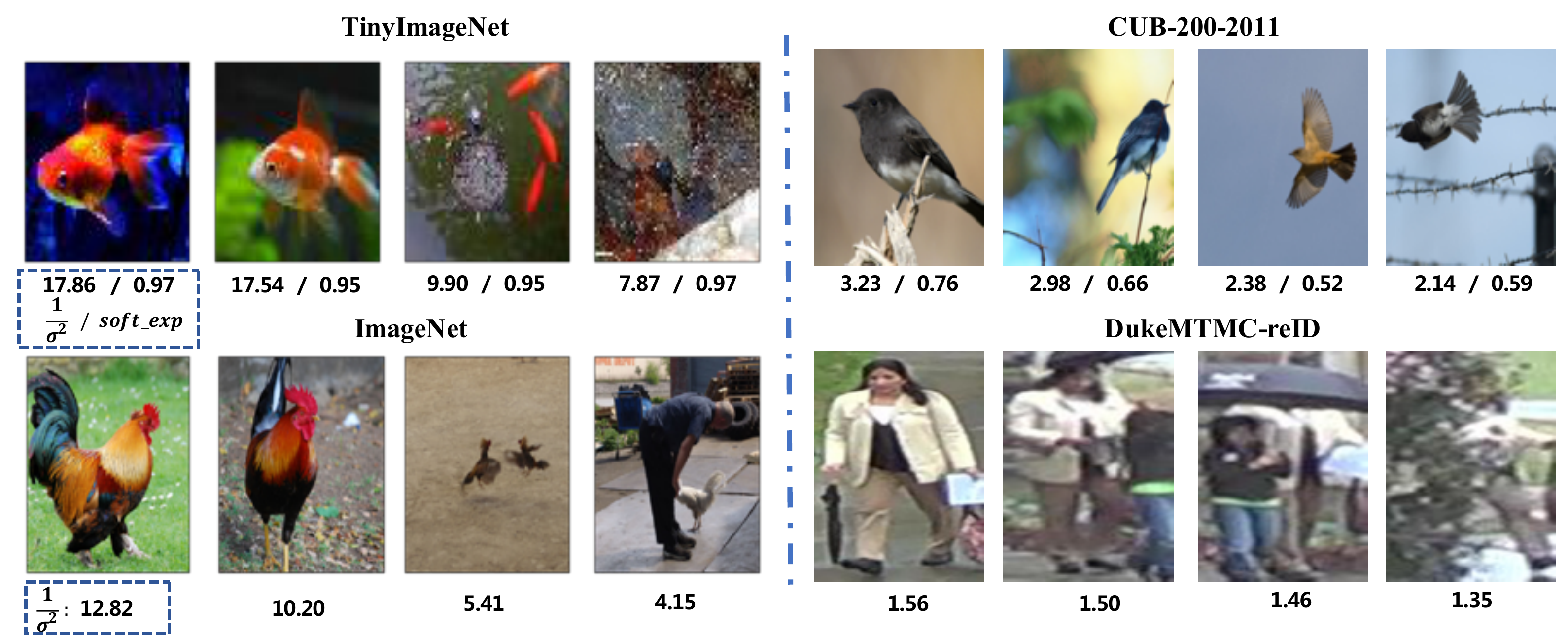}
\caption{Examples of the same class or the same person from four datasets. Generally, the larger the variance($\sigma^{2}$) is, the more difficult the sample is. Thus, PAD assigns large weights ($1/\sigma^{2}$) to the prime samples. Weights from soft-exp are not always accurate. For example, soft-exp give large weights to the right hard samples.}
\label{fig:var_vis}
\end{figure}

\noindent \textbf{How PAD Affects Distillation?}
We analyze how PAD affects the student model training in this part. From Fig.~\ref{fig:var_gap}, we realize that there is a positive correlation between the gap and the variance. From Fig.~\ref{fig:var_vis}, we observe that the quality of samples decreases with the increase of the learned variance. The above two observations are consistent with our analysis in the method part. With the help of PAD, hard samples are assigned small weights while easy samples are assigned large weights. After adaptive weighting samples by PAD, the effect of hard samples is obviously weakened, while the effect of easy ones is strengthened as the solid lines in Fig.~\ref{fig:var_gap} show. Also, PAD narrows the gaps between teacher and student, which means student trained with PAD achieves a better result in regression to the teacher than the baseline.

\begin{table}
\centering
\caption{Effect of hyper-parameters ($\alpha$ and $T$) in sample weighting baselines. ``$\uparrow$'' means better than baselines. Few experiments are better than baselines.}
\begin{tabular}{ccc|c|c|c|c|c|c}
\cline{3-9}
\multicolumn{2}{c}{}                                                        & baseline & 0.1  & 0.5  & 1    & 2    &  5  & 10   \\ \hline
\multicolumn{1}{c|}{\multirow{2}{*}{soft-exp}}  & \multicolumn{1}{c|}{TinyImageNet} & 63.1     & 61.2 & 63.2$\uparrow$ & 62.6 & 63.4$\uparrow$ & 63.6$\uparrow$ & 61.8 \\ \cline{2-9} 
\multicolumn{1}{c|}{}                           & \multicolumn{1}{c|}{CUB-200-2011}  & 58.6     & 54.9 & 58.6 & 60.1$\uparrow$ & 59.2$\uparrow$ & 55.8          & 55.3 \\ \hline
\multicolumn{1}{c|}{\multirow{2}{*}{soft-poly}} & \multicolumn{1}{c|}{TinyImageNet} & 63.1     & 62.6 & 63.0 & 62.6 & 63.5$\uparrow$ & 62.3  & 61.8  \\ \cline{2-9} 
\multicolumn{1}{c|}{}                           & \multicolumn{1}{c|}{CUB-200-2011}  & 58.6     & 59.4$\uparrow$ & 60.4$\uparrow$ & 57.9 & 57.5 & 59.5$\uparrow$          & 60.3$\uparrow$ \\ \hline
\end{tabular}
\label{table:ablation}
\end{table}

\noindent \textbf{Impact of the hyper-parameters.}\label{sec:ablation}
For sample weighting baselines, extra hyper-parameters are introduced. As shown in Table~\ref{table:ablation}, soft-weighting schemes slightly improve the performance of distillation, and good results rely heavily on the careful selection of parameters.

\subsubsection{Comparisons of different weighting schemes.} We compare different weighting schemes and draw two conclusions from Table~\ref{table:diff scheme}. First, assigning different weights to different samples is effective in distillation even using a manual way. Soft-weighting schemes perform better than the hard-discarding schemes. Second, the weights learned from the off-the-shelf PAD models can better describe the importance of samples than other schemes. Besides the quantitative indicators in Table~\ref{table:diff scheme}, we see the difference of PAD and soft-exp from the qualitative results in the first row of Fig.~\ref{fig:var_vis}.

\noindent \textbf{Distillation loss warm up.}
We also discover that variances become small at the beginning and then stable during the training process with PAD loss. This leads to an increase of weight for all samples at the beginning. To explore this phenomenon, we design a {\it {distillation loss warm up}} experiment, {\it {\it{i.e.}}}, increasing the weight of distillation loss from 0 to the default weight linearly during the initial epochs. We observe that distillation loss warm up also performs slightly better than the baseline but worse than PAD, which means both increasing the sample weight during training and assigning different samples different weights are effective techniques in distillation. PAD easily combines these two techniques and exhibits as a powerful approach to complement knowledge distillation.

\begin{table}
\begin{center}
\caption{Comparisons PAD with other weighting schemes and warm up. Top-1 accuracy on CIFAR100 and TinyImageNet. {\it R@1} on DukeMTMC-reID and CUB-200-2011. $mmAP$ on Pascal VOC07.}
\label{table:diff scheme}
\begin{tabular}{c|c|c|c|c|c}
\hline
Method                     & CIFAR100 & TinyImageNet & Duke-reID  & CUB-200 & Pascal VOC07\\ \hline
all samples equally        & 72.85    & 63.08   & 79.5          & 58.6  &45.53       \\ \hline
soft-exp                   & 73.54    & 63.64   & 80.2          & 60.1  & 46.23 \\
soft-poly                   & 73.38    & 63.50  & 79.7         & 60.4   & 46.25 \\
weights from PAD           & 73.72    & 65.54   & 80.6          &60.2   & 46.52          \\ \hline
warm up  & 73.13    & 63.94        & 80.4       & 59.0         & 46.05\\ \hline
PAD                        & 74.06    & 65.64   & 81.0          & 61.4   & 46.92      \\ \hline
\end{tabular}
\end{center}
\end{table}

\section{Conclusion}
This paper explores adaptive sample weighting in knowledge distillation, which is innovative and effective for distillation. With more attention paid to easy samples, simple weighting schemes promote the performance of knowledge distillation. Moreover, Prime-Aware Adaptive Distillation (PAD) is proposed for further improvement. PAD is seamlessly combined with existing methods to refine them by enhancing the perceived prime samples. We hope the new state-of-the-arts established by PAD can serve as a starting point for future research.

\noindent \textbf{Acknowledgements.}
This work was supported in part by the National Key Research and Development Program of China under Grant 2017YFA0700800. Thanks Xiruo Tang for her help on paper writing.

\clearpage
%
%
\bibliographystyle{splncs04}
\bibliography{egbib}
\end{document}